\documentclass{article}

\usepackage{arxiv}
\usepackage{lineno} 
\usepackage{fontspec}

\usepackage{latexsym}

\usepackage{xcolor}

\usepackage{microtype}

\usepackage{hyperref}       
\usepackage{url}            
\usepackage{booktabs}       
\usepackage{amsfonts}       
\usepackage{nicefrac}       
\usepackage{amsmath}        
\usepackage{graphicx}       
\usepackage{subcaption}
\usepackage{pgfplots}       
\pgfplotsset{compat=1.18}
\usepackage{pifont}

\usepackage{xcolor}
\definecolor{yellowbg}{RGB}{255,255,180}
\definecolor{bluebg}{RGB}{180,200,255}
\definecolor{redbg}{RGB}{255,180,180}
\newcommand{\highlight}[2]{\colorbox{#1}{\textbf{#2}}}
\usepackage{hyperref}       
\usepackage{url}            
\usepackage{booktabs}       
\usepackage{amsfonts}       
\usepackage{nicefrac}       
\usepackage{microtype}      
\usepackage{lipsum}		
\usepackage{float}
\usepackage{amsmath}
\usepackage{graphicx}
\usepackage[numbers]{natbib} 
\usepackage{doi}
\usepackage{graphicx}
\usepackage{subcaption}
\usepackage{pgfplots}
\pgfplotsset{compat=1.18}
\usepackage{pifont}

\title{When Distributions Shifts: Causal Generalization for Low-Resource Languages}

\date{} 					

\usepackage{authblk} 

\author[1,2]{Mahi Aminu Aliyu$^*$}
\author[1]{Chisom Chibuike$^*$}
\author[1]{Fatimo Adebanjo$^*$}
\author[1]{Samuel Oyeneye}
\author[1]{Omokolade Awosanya}

\affil[1]{ML Collective}
\affil[2]{Abubakar Tafawa Balewa University}

\renewcommand{\shorttitle}{\textit{arXiv} Template}

\begin{document}
\maketitle

\begin{abstract}
Machine learning models often fail under distribution shifts, a problem exacerbated in low-resource settings where limited data restricts robust generalization. Domain generalization (DG) methods address this challenge by learning representations that remain invariant across domains, frequently leveraging causal principles. In this work, we study two causal DG approaches for low-resource natural language processing. First, we apply causal data augmentation using GPT-4o-mini to generate counterfactual paraphrases for sentiment classification on the NaijaSenti Twitter corpus in Yoruba and Igbo. Second, we investigate invariant causal representation learning with the Debiasing in Aspect Review (DINER) framework for aspect-based sentiment analysis. We extend DINER to a multilingual setting by introducing Afri-SemEval, a dataset of 17 languages translated from SemEval-2014 Task. Experiments show improved robustness to unseen domains, with consistent gains from counterfactual augmentation and enhanced out-of-distribution performance from causal representation learning across multiple languages.

\end{abstract}


\section{Introduction}
\label{sec:Intro}

Machine learning models have achieved notable success in natural language processing, but their effectiveness relies on the assumption that training and test data are identically distributed. In practice, data distributions shift across time, geography, and linguistic contexts, leading to performance degradation on unseen domains. This challenge is severe in low-resource settings. In a study conducted by \citet{adebara2025sahara}, they introduced the ``Sahara" benchmark for African NLP and showed that out of 517 African languages in their benchmark, only 45 languages were present in more than one dataset, with most appearing only in basic language identification tasks. Such extreme data sparsity and fragmentation make it infeasible to rely on scale alone, amplifying the impact of distribution shifts and the need for domain generalization methods that can learn invariant representations under limited and heterogeneous data conditions \cite{eberhard2021ethnologue,joshi2020state,rescue2021translated,oyebola2023attitudes}. 

In machine learning, domain generalization (DG) addresses the challenge of performing well on unseen domains by learning representations that remain invariant across multiple source domains without access to the target distribution. This DG formulation and its associated solutions also extend to low-resource settings, where data scarcity makes robustness to distribution shifts especially important. Recent DG approaches increasingly emphasize learning stable and invariant features across domains; causal perspectives have been shown to be one effective way to formalize and motivate such invariance under domain shifts \cite{sheth2022domaingeneralizationcausal}.

\begin{figure*}[t]
\centering
    \includegraphics[width=0.9\textwidth]{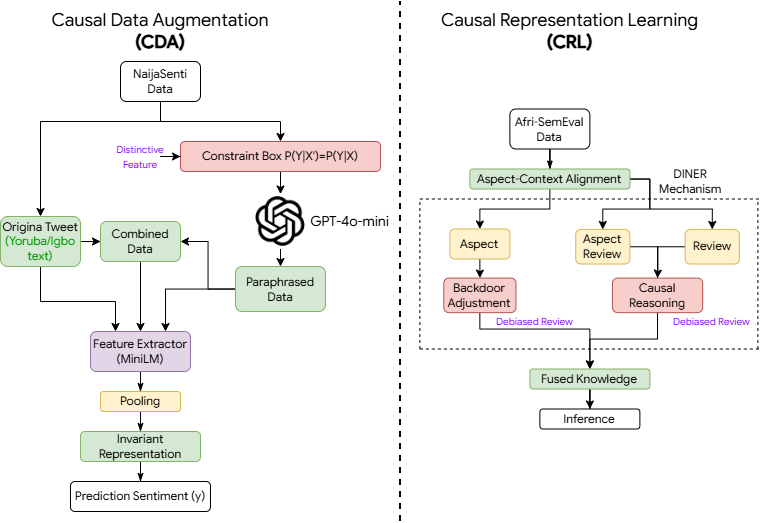}
    \caption{Overview of our unified causal framework. \highlight{redbg}{Left (CDA):} Paraphrased data is generated via \highlight{yellowbg}{\textbf{GPT-4o-mini}} under the constraint \highlight{yellowbg}{\textbf{$P(Y|X') = P(Y|X)$}}, then combined with original data for invariant representation learning. \highlight{bluebg}{Right (CRL):} The DINER mechanism applies aspect-context alignment, backdoor adjustment, and counterfactual reasoning to produce debiased sentiment predictions.}
    \label{fig:architecture}
\end{figure*}

In this study, we examine two complementary causal DG approaches for low-resource sentiment analysis: \textbf{Causal Data Augmentation (CDA)}, which generates sentiment-preserving paraphrases to simulate controlled domain shifts, and \textbf{Causal Representation Learning (CRL)}, which applies structural causal models with backdoor adjustment and counterfactual reasoning to isolate causal sentiment paths. Figure~\ref{fig:architecture} illustrates our unified framework. For CDA, we use the Yoruba and Igbo subsets of NaijaSenti \cite{muhammad2022naijasentinigeriantwittersentiment}. For CRL, we introduce \textbf{Afri-SemEval}, a new benchmark translating SemEval-2014 Task 4 \cite{marelli2014semeval} into 17 African languages.

Our contributions are: (1) a unified framework combining data-level and representation-level causal interventions for low-resource sentiment analysis; (2) Afri-SemEval, a multilingual ABSA benchmark covering 17 African languages; and (3) empirical evidence that causal mechanisms enhance generalization under distribution shifts in low-resource multilingual NLP.

In the rest of this paper, Section~\ref{sec:related_work} reviews related work on domain generalization and causal NLP. Section~\ref{sec:methodology} details our methodology for both CDA and CRL approaches. Section~\ref{sec:results} presents experimental results and analysis, and Section~\ref{sec:conclusion} concludes the paper, discusses limitations and future directions.

\section{Related Work}
\label{sec:related_work}

Domain generalization (DG) in NLP aims to improve model robustness under distribution shifts. 

Table~\ref{tab:related-work} summarizes key Domain Generalization approaches that have been explored by researchers, highlighting whether they use Causal Data Augmentation (CDA) or Causal Representation Learning (CRL) and their main contributions and limitations. 

While prior work \cite{moller2023prompt} have explored causal data augmentation (CDA) for low-resource languages, controlled sentiment-preserving paraphrasing for simulating causal domain shifts is largely unexplored. Additionally, causal representation learning (CRL) has mostly been applied to high-resource languages.

Our study explores two causal frameworks for low-resource African languages; CDA and CRL. Unlike previous works, we explored CDA via sentiment-preserving paraphrases in Yoruba and Igbo. For CRL, we introduced Afri-SemEval, a multilingual ABSA dataset covering 17 African languages, enabling evaluation of domain generalization under distribution shifts in low-resource multilingual settings.

\begin{table*}[t]
\centering
\small
\begin{tabular}
{p{2.2cm}p{2.3cm}p{2.5cm}p{2cm}p{4.5cm}}
\toprule
\textbf{Authors} & \textbf{Technique} & \textbf{Research Goal} & \textbf{CDA/CRL} & \textbf{Key Contribution} \\
\midrule
\citet{wu2025m} & Multilingual ABSA Benchmark (M-ABSA) & Multilingual aspect-based sentiment analysis & CRL & Addressed data scarcity via benchmark construction and fine-tuning of generative models. \\
\midrule
\citet{muhammad2022naijasentinigeriantwittersentiment}; \citet{muhammad2023afrisenti} & Dataset Creation (NaijaSenti, AfriSenti) & Sentiment analysis for African languages & CDA & Pioneered African language sentiment datasets but limited to thousands of examples. \\
\midrule
\citet{ibrahim2025deep}; \citet{maqsood2025towards} & Low-resource Benchmark Creation & Sentiment analysis for Hausa and Urdu & CRL & Introduced language-specific benchmarks that improve coverage but not address robustness in domain shifts. \\
\midrule
\citet{kaushik2020learning} & Counterfactual Augmentation & Reduce spurious feature sensitivity & CDA & Showed 40--60\% reduction in spurious correlations via counterfactuals. \\
\midrule
\citet{vsmid2025improving} & Constrained Decoding & Syntactic validity in cross-lingual generation & CRL & Enforced syntactic constraints during generation without addressing semantic or causal robustness. \\

\midrule
\citet{balashankar2023improving} & Active Learning + Counterfactual Generation & Efficient robustness with minimal annotation & CDA & Achieved 18--20\% robustness gains with 10\% human annotation. \\
\midrule
\citet{moller2023prompt} & GPT-4 Data Augmentation & Low-resource classification tasks & CDA & Noted quality degradation for low-resource languages. \\
\midrule
\citet{wu2024diner} & DINER  & Debias ABSA via causal inference & CRL & Backdoor adjustment and counterfactual reasoning to English ABSA. \\
\bottomrule
\end{tabular}
\caption{Overview of domain generalization approaches relevant to low-resource NLP. Techniques are \highlight{redbg}{\textbf{categorized}} by \highlight{bluebg}{\textbf{authors}}, \highlight{yellowbg}{\textbf{method}}, \highlight{redbg}{\textbf{research goal}}, and \highlight{bluebg}{type of \textbf{causal intervention}}: Causal Data Augmentation (CDA) or Causal Representation Learning (CRL). Key contributions highlight each method's main achievements and limitations.}
\label{tab:related-work}
\end{table*}
\section{Methodology}
\label{sec:methodology}
This section presents the methodological design adopted in this study, which comprises of two experimental frameworks: (1)Invariance via Causal Data Augmentation and (2) Invariance via Causal Representation Learning.

\subsection{Invariance via Causal Data Augmentation}
\label{sec:causal_method_data_aug}

Here, we investigate \textbf{causal data augmentation} as a strategy for improving sentiment classification in low-resource African languages through representation invariance. The approach is grounded in the principles of causal representation learning: if the sentiment label $y$ depends on the semantic content of a tweet $x$, then augmentations that perturb only non-causal linguistic features (e.g., surface form, style, or syntax) should preserve the conditional probability distribution $P(y|x)$.

We used the Yoruba and Igbo subsets of the \textit{NaijaSenti} corpus\cite{muhammad2022naijasentinigeriantwittersentiment}, a multilingual sentiment twitter dataset covering four Nigerian languages: Yoruba, Igbo, Hausa, and Nigerian Pidgin. Each instance is a tweet–label pair $(x_i, y_i)$, where $x_i$ represents the tweet and $y_i \in \{\text{positive}, \text{negative}, \text{neutral}\}$ is its sentiment label.  

\subsubsection{Experiment Procedure}

\paragraph{Notation.} Let $\mathcal{X}$ denote the input text space and $\mathcal{Y}$ the sentiment label space. Let $\mathcal{D}_1 = \{(x_i, y_i)\}_{i=1}^{N}$ be the original sentiment dataset sampled from a joint distribution $P(X,Y)$, where $x_i \in \mathcal{X}$ and $y_i \in \mathcal{Y}$.
Let $f:\mathcal{X}\rightarrow\mathcal{X}'$ be a paraphrasing function producing $x' = f(x)$.
We denote the paraphrased dataset as $\mathcal{D}_2 = \{(x'_i, y_i)\}_{i=1}^{N}$, and the combination of the paraphrased and original dataset as $\mathcal{D}_{\text{comb}} = \mathcal{D}_1 \cup \mathcal{D}_2$.

\paragraph{Data Augmentation via Paraphrasing.}
To simulate controlled domain shifts while preserving label semantics, we construct our paraphrased dataset $\mathcal{D}_2$ using a sentiment-preserving augmentation function $f:\mathcal{X}\rightarrow\mathcal{X}'$ such that $P(Y \mid X') = P(Y \mid X)$. The paraphrases are generated using the \textbf{GPT-4o-mini} model~\cite{openai_gpt4o_mini_2024} with a carefully designed prompt that explicitly enforces sentiment and semantic preservation while introducing lexical and syntactic variation. The prompt template is shown below:
\begin{quote}
\textit{``You have been provided with a \{language\} tweet. Paraphrase `\{tweet\}' in a way that preserves its sentiment and meaning but changes its wording. Return only the paraphrased tweet."}
\end{quote}
This process yields a paraphrased dataset $\mathcal{D}_2$ that is semantically aligned with the original dataset $\mathcal{D}_1$. We observed that GPT-4o-mini preserves sentiment polarity while introducing lexical and syntactic diversity which retains the underlying sentiment and lies in the same semantic neighborhood. The illustration in Table~\ref{tab:cda_examples} are examples of original tweets and their corresponding paraphrases (`Y" denotes Yoruba, `I" denotes Igbo). 

\begin{table}[ht]
\centering
\small
\begin{tabular}{p{2.4cm}p{2.8cm}c}
\toprule
\textbf{Original Tweet ($X$)} & \textbf{Paraphrased Tweet ($X'$) (GPT-4o-mini)} & \textbf{Label ($Y$)} \\
\midrule
\textit{Y: }``Oúnjẹ yìí dára púpọ̀, mo fẹ́ràn rẹ̀ gan-an'' & ``Oúnjẹ náà dùn púpọ̀, ó wù mí gidigidi'' & \textcolor{green!60!black}{positive} \\
\midrule
\textit{I: }``Ahịa a jọrọ njọ, ọ gaghị m azụta ya ọzọ'' &  ``Ngwa ahịa a adịghị mma, agaghị m alọghachikwuru ya'' & \textcolor{red}{negative} \\
\midrule
\textit{Y: }``Ayé ilá ilá ikàn wọ ẹ̀wù ẹ̀jẹ̀ sọ ọrún ọ̀pá gbédìí fáyé'' &  ``Ayé ilá ilá ikàn wọ aṣọ ẹ̀jẹ̀ sọ ọrún ọ̀pá gbédìí fáyé'' & \textcolor{red}{negative} \\
\bottomrule
\end{tabular}
\caption{Examples of sentiment-preserving paraphrases generated by GPT-4o-mini for Yoruba and Igbo tweets.}
\label{tab:cda_examples}
\end{table}

One of the main challenges in guiding LLMs to generate diverse variants while maintaining key information arises for proverbial and sarcastic tweets, where sentiment is often encoded through more nuanced expressions. In such cases, the model’s ability to retain variations of the original sentiment without diluting the underlying proverbial meaning or sarcasm is limited. To mitigate this issue, we employ a human annotation and correction mechanism for a subset of tweets exhibiting these challenges.

Our resulting output exhibits a shift in the marginal input distribution:
\begin{equation}
\begin{split}
    P_{\mathcal{D}_2}(X') &\neq P_{\mathcal{D}_1}(X), \\
    P_{\mathcal{D}_2}(Y \mid X') &\approx P_{\mathcal{D}_1}(Y \mid X).
\end{split}
\end{equation}


\begin{figure}[htbp]
\centering
    \includegraphics[width=0.5\textwidth]{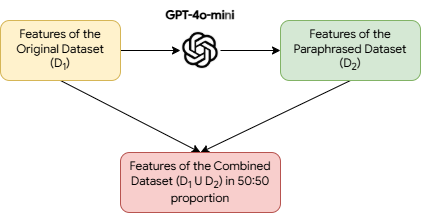}
    \caption{Flow of $\mathcal{D}_1$, $\mathcal{D}_2$, and $\mathcal{D}_{\text{comb}}$}
    \label{fig:paraphrase}
\end{figure}

\begin{figure}[htbp]
\centering
    \includegraphics[width=0.7\textwidth]{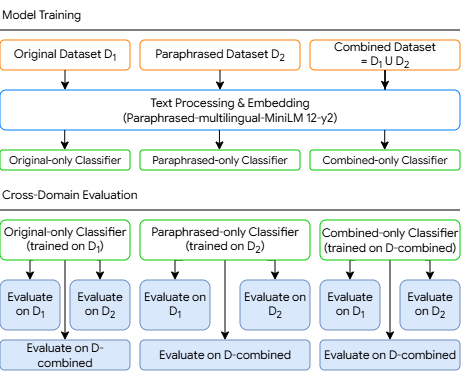}
    \caption{Flow of the experiment procedure used in Invariance via Causal Data Augmentation.}
    \label{fig:proceed}
\end{figure}

For each language subset, we train three logistic regression classifiers: \textbf{Original-only}, trained on $\mathcal{D}_1$; \textbf{Paraphrased-only}, trained on $\mathcal{D}_2$; and \textbf{Combined}, trained on $\mathcal{D}_{\text{comb}}$ with a balanced 50–50 mix (this 50:50 mixing strategy is used to avoid bias stemming from over-representation of either original or paraphrased samples during training). We choose logistic regression for its interpretability and well-understood optimization properties, allowing us to isolate the effects of data augmentation and distributional shift without confounding factors introduced by model complexity. To assess robustness under the introduced distribution shift, we adopt a cross-domain evaluation protocol. Each model is evaluated on $\mathcal{D}_1$, $\mathcal{D}_2$, and $\mathcal{D}_{\text{comb}}$, enabling systematic comparison of in-domain and cross-domain generalization performance. See Figure~\ref{fig:proceed}.

Each classifier $h_\theta$ is trained by minimizing the empirical cross-entropy loss:
\begin{equation}
    \hat{\mathcal{L}}(\theta) = \frac{1}{|\mathcal{D}_{\text{train}}|} 
    \sum_{(x,y)\in\mathcal{D}_{\text{train}}}
    \ell\big(h_\theta(g_\psi(\phi(x))), y\big),
\end{equation}
where $\phi$ denotes text preprocessing (tokenization, normalization, stopword removal), and $g_\psi$ is a fixed sentence embedding function implemented using the \textit{paraphrase-multilingual-MiniLM-L12-v2} model~\citep{reimers-2019-sentence-bert}. Our model performance was measured using ``accuracy" for classification tasks.
This experimental design enables a principled empirical test of our causal invariance hypothesis, by evaluating whether models trained with label-preserving augmentations improves generalization under distributional shift in low-resource settings.

\subsection{Invariance via Causal Representation Learning}

Invariance via Causal Representation Learning (CRL) during inference is one of the causal domain generalization techniques we explore in low-resource settings. To achieve this, we adopt the \textbf{DINER}: Debiasing Aspect-based Sentiment Analysis with Multi-variable
Causal Inference \cite{wu2024diner}, a debiasing approach for the Aspect-Based Sentiment Analysis (ABSA) task. We discuss our methodology in what follows.

\subsubsection{Debiasing in Aspect Review (DINER)}

Recent studies have proposed several debiasing strategies, including argumentation-based methods \cite{wei2019eda}, reweighting-based training \cite{schuster2019debiasingfactverificationmodels}, and causal inference-based approaches \cite{niu2021counterfactual, liu2021towards}. Among these, causal inference has attracted significant interest for its solid theoretical foundation and minimal disruption to existing learning paradigms. See Section~\ref{sec:appendix} for more details on debiasing in aspect review using DINER.

\subsubsection{Data}

\begin{figure}[htbp]
\centering
    \includegraphics[width=0.3\textwidth]{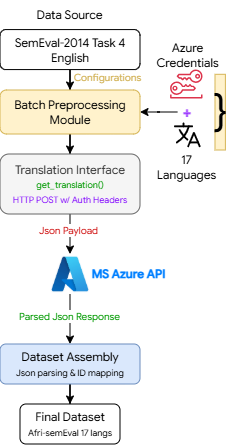}
    \caption{Afri-SemEval dataset creation pipeline showing the translation workflow for SemEval-2014 to 17 African languages using Azure API.}
    \label{fig:afrisemeval_pipeline}
\end{figure}

We introduce \textbf{Afri-SemEval}, a multilingual dataset created by translating the SemEval 2014 dataset into 17 African languages to support research on domain generalization in low-resource contexts. Figure~\ref{fig:afrisemeval_pipeline} illustrates our dataset creation pipeline. We leverage the Microsoft Azure Translator API for one-to-many translation, where English reviews are batch-processed with configurations settings to comply with API constraints and translated simultaneously into all target languages. The pipeline parses JSON responses, maps translations to source review IDs, and applies a semantic similarity heuristic to align aspect terms across languages. The dataset spans diverse languages from multiple African language families, as shown in Table~\ref{tab:languages}.

\begin{table}[ht]
  \centering
  \small
  \setlength{\tabcolsep}{3pt}
  \begin{tabular}{clc|clc}
    \toprule
    \textbf{S/N} & \textbf{Language} & \textbf{Code} & \textbf{S/N} & \textbf{Language} & \textbf{Code} \\
    \midrule
    1 & Afrikaans & af & 10 & Kinyarwanda & rw \\
    2 & Amharic & am & 11 & Shona & sn \\
    3 & Hausa & ha & 12 & Somali & so \\
    4 & Igbo & ig & 13 & S. Sotho & st \\
    5 & Lingala & ln & 14 & Tswana & tn \\
    6 & Luganda & lug & 15 & Xhosa & xh \\
    7 & N. Sotho & nso & 16 & Yoruba & yo \\
    8 & Chichewa & nya & 17 & Zulu & zu \\
    9 & Kirundi & run &    &         &    \\
    \bottomrule
  \end{tabular}
  \caption{List of 17 African supported languages}
  \label{tab:languages}
\end{table}

Each translated instance maintains the original SemEval-2014 format, comprising review text, target aspect, and sentiment polarity (positive, negative, or neutral). The translation process involved ensuring aspect alignment, polarity preservation, and accounting for linguistic and cultural nuances in sentiment expression.

A major challenge in adapting DINER to African languages lies in aspect term alignment. The original framework assumes a direct mapping between aspect terms and their positions in the review, but translation often disrupts this due to word order variation, morphological complexity (e.g., noun classes in Bantu languages), and differences in expression length (multiword to single-word mappings, and vice versa). 

We address this with an alignment heuristic based on semantic similarity. Embeddings are computed for translated aspect terms and all n-grams in the review. The n-gram with the highest similarity to the aspect term is then selected as its aligned position for DINER processing. While effective for single-token aspects, this heuristic remains challenging for multi-token and non-contiguous aspect terms.

\section{Experiment and Result}
\label{sec:results}

\subsection{Invariance via Causal Data Augmentation}
\label{sec:causal_data_results}

All experiments were conducted on the Yoruba and Igbo subsets of the \textit{NaijaSenti} dataset\cite{muhammad2022naijasentinigeriantwittersentiment}, following the causal data augmentation setup described in Section~\ref{sec:causal_method_data_aug}. For each language, three logistic regression classifiers were trained separately on: (i) the original dataset $\mathcal{D}_1$, (ii) the paraphrased dataset $\mathcal{D}_2$ generated using GPT-4o-mini, and (iii) the combined dataset $\mathcal{D}_1 \cup \mathcal{D}_2$. Evaluation was performed across all possible test configurations; original, paraphrased, and combined to assess within-domain and cross-domain generalization.

The core objective is to test the \textbf{causal invariance hypothesis}, which posits that augmentations preserving causal semantics ($P(y|x') = P(y|x)$) encourage models to learn representations invariant to superficial linguistic variations. Formally, we aim to minimize the expected difference in accuracy across domains:
\begin{equation}
    \underset{\theta}{\text{argmin}} \;
    \mathbb{E}_{d_i, d_j \in \{\mathcal{D}_1, \mathcal{D}_2\}}
    \big[\, |\text{Acc}_{d_i}(h_\theta) - \text{Acc}_{d_j}(h_\theta)| \,\big],
\end{equation}
where $h_\theta$ denotes the trained classifier.

\subsubsection{Results on Yoruba} 

Tables~\ref{tab:yoruba_val_results} and~\ref{tab:yoruba_test_results} summarize the validation and test accuracies for Yoruba, respectively. The combined configuration consistently yields higher accuracy across domains, confirming that integrating both original and paraphrased data enhances generalization. 

\begin{table}[htbp]
\centering
\small
\begin{tabular}{llll}
\toprule
\textbf{Train Dataset} & \textbf{Test Dataset} & \textbf{Accuracy} \\
\midrule
Original & Original & 0.6787 \\
Original & Paraphrased & 0.6602 \\
Original & Combined & 0.6925 \\
Paraphrased & Original & 0.6584 \\
Paraphrased & Paraphrased & 0.6787 \\
Paraphrased & Combined & 0.6907 \\
\textbf{Combined} &\textbf{ Original }& \textbf{0.6999} \\
Combined & Paraphrased & 0.6898 \\
Combined & Combined & 0.6934 \\
\bottomrule
\end{tabular}
\caption{Validation results on Yoruba subset.}
\label{tab:yoruba_val_results}
\end{table}

\begin{table}[htbp]
\centering
\small
\begin{tabular}{lll}
\toprule
\textbf{Train Dataset} & \textbf{Test Dataset} & \textbf{Accuracy} \\
\midrule
Original & Test Set & 0.6799 \\
Paraphrased & Test Set & 0.6747 \\
\textbf{Combined} & \textbf{Test Set} & \textbf{0.6892} \\
\bottomrule
\end{tabular}
\caption{Test results on Yoruba subset.}
\label{tab:yoruba_test_results}
\end{table}

Across the Yoruba results, the model trained on the combined dataset achieved the highest test accuracy of \textbf{68.92\%}, compared to \textbf{67.99\%} for the original-only model and \textbf{67.47\%} for the paraphrased-only model. This demonstrates that exposure to both natural and paraphrased data improves robustness to distributional variation. Moreover, the smaller cross-domain deviation in validation accuracies ($|\text{Acc}_{\text{orig}} - \text{Acc}_{\text{para}}| \approx 0.01$) further confirms increased invariance to paraphrastic shifts.

\subsubsection{Results on Igbo}
Tables~\ref{tab:igbo_val_results} and~\ref{tab:igbo_test_results} present the corresponding results for the Igbo subset. The general trend in this case, does not mirror Yoruba. While the combined dataset achieves stable performance across domains, improvements are more pronounced on our original data due to a larger semantic drift between original and paraphrased data.

\begin{table}[htbp]
\centering
\small
\begin{tabular}{llll}
\toprule
\textbf{Train Dataset} & \textbf{Test Dataset} & \textbf{Accuracy} \\
\midrule
\textbf{Original }& \textbf{Original} & \textbf{0.7590} \\
Original & Paraphrased & 0.6429 \\
Original & Combined & 0.7230 \\
Paraphrased & Original & 0.6922 \\
Paraphrased & Paraphrased & 0.6834 \\
Paraphrased & Combined & 0.7502 \\
Combined & Original & 0.7370 \\
Combined & Paraphrased & 0.6895 \\
Combined & Combined & 0.7326 \\
\bottomrule
\end{tabular}
\caption{Validation results on Igbo subset.}
\label{tab:igbo_val_results}
\end{table}

\begin{table}[htbp]
\centering
\small
\begin{tabular}{lll}
\toprule
\textbf{Train Dataset} & \textbf{Test Dataset} & \textbf{Accuracy} \\
\midrule
\textbf{Original} &\textbf{ Test Set} & \textbf{0.7283} \\
Paraphrased & Test Set & 0.7084 \\
Combined & Test Set & 0.7176 \\
\bottomrule
\end{tabular}
\caption{Test results on Igbo subset.}
\label{tab:igbo_test_results}
\end{table}

While the Igbo combined model achieved a slightly lower test accuracy (71.76\%) compared to the original-only model (72.83\%), it exhibited more consistent validation performance across paraphrastic and combined domains, indicating reduced overfitting to surface lexical cues. This suggests that the degree of causal alignment between paraphrased and original text affects the strength of invariance gained.

\vspace{0.6cm}
Overall, our results confirm that causal-preserving data augmentation yields measurable improvements in cross-domain robustness. In Yoruba, where paraphrases were semantically well-aligned, the combined model achieved the best generalization with the smallest invariance gap $\Delta_{\text{inv}}$. In Igbo, where paraphrases had a larger distributional shift, the combined model still maintained competitive performance, with smaller gains. These findings suggests that training on both original and paraphrased data enhances model invariance to superficial linguistic variability. The modest but consistent improvements show the potential of causal data augmentation as a lightweight strategy for generalization in low-resource sentiment classification.

\subsection{Invariance via Causal Representation Learning}

For our experiments, we evaluated three transformer-based models: XLM-R~\cite{conneau2019unsupervised}, Afro-XLMR Large~\cite{alabi-etal-2022-adapting}, and Afro-XLMR Large-76L~\cite{adelani2023sib200}. For each model, we compare two training configurations: CT=False (standard training with concatenated aspect and review) and CT=True (causal training with separate aspect and review processing).

All models were fine-tuned with learning rate 2e-5, batch size 16, maximum sequence length 128, 10 epochs, weight decay 0.01, and AdamW optimizer. For CT=True models, we implemented DINER's causal inference components including backdoor adjustment ($\tau=0.1$), counterfactual reasoning, and TIE calculation with ensemble fusion.

We use accuracy as metrics to access the performance of our approach. To assess out-of-distribution (OOD) generalization, we employ cross-language evaluation and perturbation tests including RevTgt (reversing target aspect polarity), RevNon (reversing non-target aspect polarity), and AddDiff (adding aspects with different polarity).

\subsubsection{Results}

\begin{figure}[!t]
  \centering
  \includegraphics[width=0.7\columnwidth]{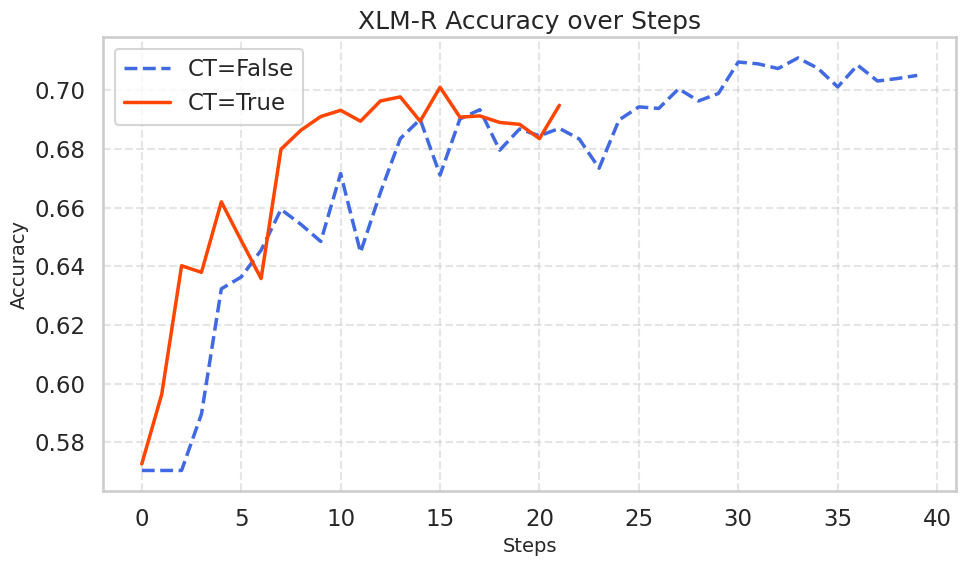}
  \caption{XLM-R accuracy comparison showing convergence patterns for CT=False and CT=True configurations.}
  \label{fig:xlmr_accuracy}
\end{figure}

\begin{figure}[!t]
  \centering
  \includegraphics[width=0.7\columnwidth]{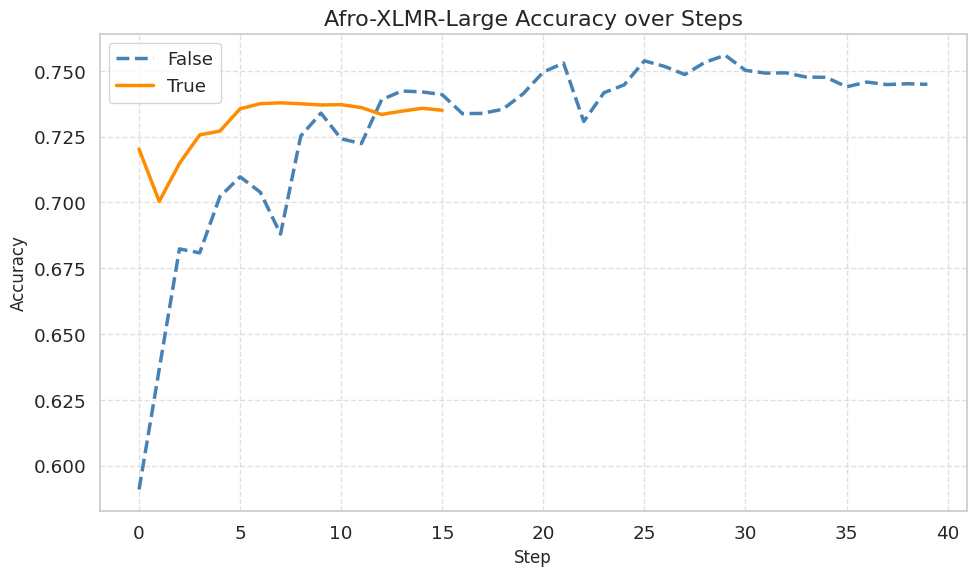}
  \caption{Afro-XLMR-large accuracy comparison showing convergence patterns for CT=False and CT=True configurations.}
  \label{fig:afro_xlmr_large}
\end{figure}

\begin{figure}[!t]
  \centering
  \includegraphics[width=0.7\columnwidth]{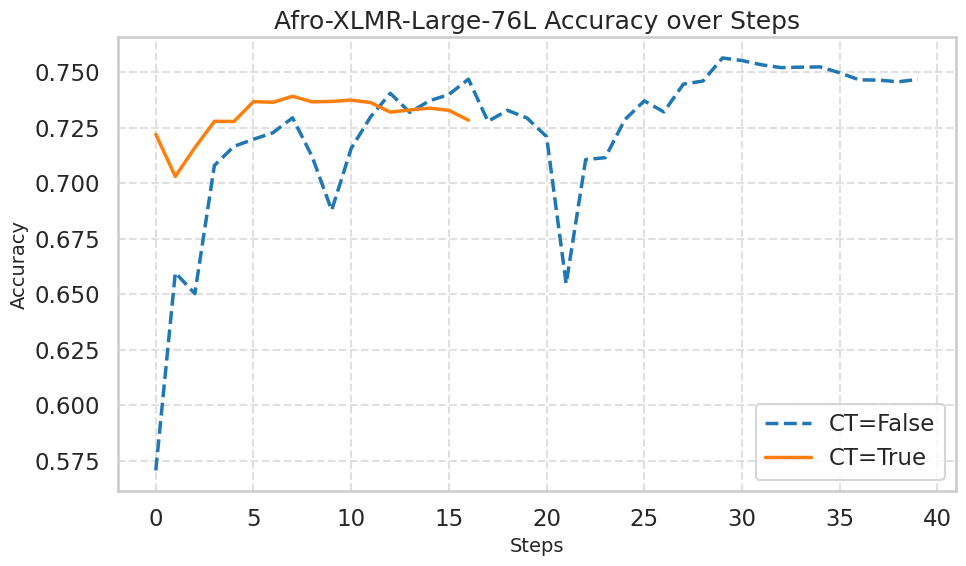}
  \caption{Afro-XLMR-large-76L accuracy comparison showing convergence patterns for CT=False and CT=True configurations.}
  \label{fig:afro_xlmr_large_76l}
\end{figure}

A key finding is that causal training (CT=True) typically converges faster across multiple transformer models. As shown in Figures~\ref{fig:xlmr_accuracy}, \ref{fig:afro_xlmr_large}, and~\ref{fig:afro_xlmr_large_76l}, models trained with causal inference require fewer steps to reach comparable performance. This efficiency advantage is particularly valuable in resource-constrained contexts, enabling faster training cycles and reduced computational demands.

Performance varies across African languages, reflecting both linguistic differences and translation quality.  Table~\ref{tab:performance} presents the performance results across five low-resource African languages for three transformer-based models under both standard training (CT=False) and causal training (CT=True) configurations. Afrikaans demonstrates the highest overall performance, likely due to its similarity to English. Amharic shows the lowest performance, reflecting script and morphological complexity. Hausa, Igbo, and Yoruba exhibit moderate performance with varying gains under causal training.

\begin{table*}[!t]
    \small
    \centering
    \begin{tabular}{lcccccc}
        \toprule
        \textbf{Model} & \textbf{Setting} & \textbf{Afrikaans} & \textbf{Amharic} & \textbf{Hausa} & \textbf{Igbo} & \textbf{Yoruba} \\
        \midrule
        XLM-R & CT=False & 78.2 & 62.1 & 70.4 & 71.5 & 72.3 \\
        XLM-R & CT=True & \textbf{79.6} & \textbf{63.5} & \textbf{72.8} & \textbf{74.2} & \textbf{75.1} \\
        \addlinespace
        Afro-XLMR Large & CT=False & 80.4 & 64.7 & 72.5 & 73.7 & 74.6 \\
        Afro-XLMR Large & CT=True & \textbf{81.1} & \textbf{65.2} & \textbf{74.1} & \textbf{74.9} & \textbf{75.8} \\
        \addlinespace
        Afro-XLMR Large-76L & CT=False & 81.6 & 65.3 & 73.2 & 74.5 & 75.4 \\
        Afro-XLMR Large-76L & CT=True & \textbf{82.5} & \textbf{66.1} & \textbf{74.9} & \textbf{76.1} & \textbf{77.2} \\
        \bottomrule
    \end{tabular}
    \caption{Performance on test set for 5 low-resource languages (Accuracy \%)}
    \label{tab:performance}
\end{table*}
The theoretical advantage of causal models in OOD settings shows mixed results in practice. Table~\ref{tab:ood} shows the OOD generalization results across the same five languages, demonstrating varying effectiveness of causal training across different model architectures. The results reveals that XLM-R generally benefits from CT=True for Afrikaans, Igbo, and Yoruba, while AfroXLM-R Large shows inconsistent patterns. These inconsistencies likely stem from translation noise affecting causal variable alignment and varying degrees of distributional shift across languages.

\begin{table*}[!ht]
    \centering
    \small
    \begin{tabular}{lcccccc}
        \toprule
        \textbf{Model} & \textbf{Setting} & \textbf{Afrikaans} & \textbf{Amharic} & \textbf{Hausa} & \textbf{Igbo} & \textbf{Yoruba} \\
        \midrule
        XLM-R & CT=False & 52.4 & \textbf{48.3} & \textbf{49.2} & 47.6 & 48.5 \\
        XLM-R & CT=True & \textbf{55.8} & 47.9 & 48.7 & \textbf{51.2} & \textbf{52.1} \\
        \addlinespace
        Afro-XLMR Large & CT=False & \textbf{54.7} & \textbf{50.1} & \textbf{51.3} & \textbf{49.8} & 49.6 \\
        Afro-XLMR Large & CT=True & 53.5 & 48.7 & 49.5 & 48.4 & \textbf{50.3} \\
        \addlinespace
        Afro-XLMR Large-76L & CT=False & \textbf{41.02}  & \textbf{41.14} & \textbf{41.28} & \textbf{37.79} & \textbf{39.41} \\
         Afro-XLMR Large-76L & CT=True & 34.52 & 34.92 & 35.13 & 30.89 & 33.41 \\
        \bottomrule
    \end{tabular}
     \caption{Performance in OOD settings (Accuracy \%)}
    \label{tab:ood}
\end{table*}

In section ~\ref{sec:case_study}, we further discussed some examples and case studies that
demonstrates the robustness improvements from the causal training in our CRL experiments. See table  ~\ref{tab:case_studies}.
\section{Conclusion and Future Work}
\label{sec:conclusion}

In this study, we investigated two causal domain generalization techniques for low-resource African languages: Causal Data Augmentation (CDA) and Invariant Causal Representation Learning (ICRL) via the DINER framework.

Our findings confirm that both approaches enhance model robustness against distribution shifts. CDA experiments on NaijaSenti showed that augmenting with semantically equivalent paraphrases improves cross-domain generalization, particularly in the Yoruba dataset where paraphrase quality was high. In contrast, Igbo results highlight that success depends on semantic fidelity, as distributional drift can limit gains. Applying DINER to the Afri-SemEval dataset demonstrated that causal training accelerates convergence and improves in-domain accuracy across 17 languages, while out-of-distribution generalization varied. This variability reveals a dual-shift problem, where models face both linguistic domain shifts and artifacts introduced by translation noise and aspect misalignment.

Future work should address three directions: developing robust aspect alignment methods for typologically diverse languages, improving LLM-generated paraphrase quality through controlled generation or human-in-the-loop validation, and designing domain generalization methods robust to both natural linguistic variation and cross-lingual transfer noise. A hybrid approach combining CDA-enhanced pretraining with representation-level debiasing may be effective.

\vspace{0.5em} \noindent \textbf{Ethical Considerations}
This work focuses on low-resource languages, which are historically underrepresented in NLP. While our dataset and models aim to improve language coverage and model generalization in distribution shift settings, they may still reflect cultural or social biases present in the source data.

We only use publicly available text and synthetic paraphrases, avoiding private or sensitive information. Finally, generated paraphrases should respect cultural norms to avoid offensive or misleading outputs, ensuring responsible and ethical use of the resources we provide.

\vspace{0.5em} \noindent \textbf{Limitations} 
While this study demonstrates the effectiveness of causal domain generalization techniques in low-resource NLP, several limitations remain. First, the Afri-SemEval dataset relies on machine translation, which may introduce artifacts and fail to capture nuances of native African language expressions. Second, our experiments focus on sentiment analysis, and the generalizability of these findings to other NLP tasks remains unexamined. Third, the computational cost of counterfactual data augmentation using large language models may limit scalability in resource-constrained settings. Finally, our evaluation covers only 17 African languages, and broader linguistic coverage would strengthen the conclusions.

\vspace{0.5em} \noindent \textbf{Acknowledgment}: 

We are grateful to ML Collective for generously providing the computational resources required for this study. The success of our extensive experiments, in particular, the resource-intensive large language model (LLM) fine-tuning and the multi-language DINER training, was directly enabled by their dedicated compute support.

\bibliographystyle{unsrtnat}

\begin{thebibliography}{29}
\providecommand{\natexlab}[1]{#1}
\providecommand{\url}[1]{\texttt{#1}}
\expandafter\ifx\csname urlstyle\endcsname\relax
  \providecommand{\doi}[1]{doi: #1}\else
  \providecommand{\doi}{doi: \begingroup \urlstyle{rm}\Url}\fi

\bibitem[Adebara et~al.(2025)Adebara, Toyin, Ghebremichael, Elmadany, and Abdul-Mageed]{adebara2025sahara}
Ife Adebara, Hawau~Olamide Toyin, Nahom~Tesfu Ghebremichael, AbdelRahim~A. Elmadany, and Muhammad Abdul-Mageed.
\newblock Where are we? evaluating {LLM} performance on {A}frican languages.
\newblock In \emph{Proceedings of the 63rd Annual Meeting of the Association for Computational Linguistics (Volume 1: Long Papers)}, pages 32704--32731, Vienna, Austria, 2025. Association for Computational Linguistics.

\bibitem[Eberhard et~al.(2021)Eberhard, Simons, and Fennig]{eberhard2021ethnologue}
David~M. Eberhard, Gary~F. Simons, and Charles~D. Fennig.
\newblock \emph{Ethnologue: Languages of the World}.
\newblock SIL International, Dallas, Texas, 24 edition, 2021.
\newblock URL \url{http://www.ethnologue.com}.

\bibitem[Joshi et~al.(2020)Joshi, Santy, Budhiraja, Bali, and Choudhury]{joshi2020state}
Pratik Joshi, Sebastin Santy, Amar Budhiraja, Kalika Bali, and Monojit Choudhury.
\newblock The state and fate of linguistic diversity and inclusion in the nlp world.
\newblock In \emph{Proceedings of the 58th Annual Meeting of the Association for Computational Linguistics}, pages 6282--6293, 2020.

\bibitem[ResCue and Agbozo(2021)]{rescue2021translated}
Elvis ResCue and G.~Edzordzi Agbozo.
\newblock Creating translated interfaces: The representations of {A}frican languages and cultures in digital media.
\newblock In Leketi Makalela and Goodith White, editors, \emph{Rethinking Language Use in Digital Africa: Technology and Communication in Sub-Saharan Africa}, pages 51--72. Multilingual Matters, Bristol, UK, 2021.

\bibitem[Oyebola and Ugwuanyi(2023)]{oyebola2023attitudes}
Folajimi Oyebola and Kingsley Ugwuanyi.
\newblock Attitudes of {N}igerians towards {BBC} {P}idgin: A preliminary study.
\newblock \emph{Language Matters: Studies in the Languages of Africa}, 54\penalty0 (1):\penalty0 78--101, 2023.
\newblock \doi{10.1080/10228195.2023.2203509}.

\bibitem[Sheth et~al.(2022)Sheth, Moraffah, Candan, Raglin, and Liu]{sheth2022domaingeneralizationcausal}
Paras Sheth, Raha Moraffah, K.~Selçuk Candan, Adrienne Raglin, and Huan Liu.
\newblock Domain generalization -- a causal perspective, 2022.
\newblock URL \url{https://arxiv.org/abs/2209.15177}.

\bibitem[Muhammad et~al.(2022)Muhammad, Adelani, Ruder, Ahmad, Abdulmumin, Bello, Choudhury, Emezue, Abdullahi, Aremu, Jeorge, and Brazdil]{muhammad2022naijasentinigeriantwittersentiment}
Shamsuddeen~Hassan Muhammad, David~Ifeoluwa Adelani, Sebastian Ruder, Ibrahim~Said Ahmad, Idris Abdulmumin, Bello~Shehu Bello, Monojit Choudhury, Chris~Chinenye Emezue, Saheed~Salahudeen Abdullahi, Anuoluwapo Aremu, Alipio Jeorge, and Pavel Brazdil.
\newblock Naijasenti: A nigerian twitter sentiment corpus for multilingual sentiment analysis, 2022.
\newblock URL \url{https://arxiv.org/abs/2201.08277}.

\bibitem[Marelli et~al.(2014)Marelli, Bentivogli, Baroni, Bernardi, Menini, and Zamparelli]{marelli2014semeval}
Marco Marelli, Luisa Bentivogli, Marco Baroni, Raffaella Bernardi, Stefano Menini, and Roberto Zamparelli.
\newblock Semeval-2014 task 1: Evaluation of compositional distributional semantic models on full sentences through semantic relatedness and textual entailment.
\newblock In \emph{Proceedings of the 8th international workshop on semantic evaluation (SemEval 2014)}, pages 1--8, 2014.

\bibitem[M{\o}ller et~al.(2024)M{\o}ller, Pera, Dalsgaard, and Aiello]{moller2023prompt}
Anders~Giovanni M{\o}ller, Arianna Pera, Jacob Dalsgaard, and Luca Aiello.
\newblock The parrot dilemma: Human-labeled vs. llm-augmented data in classification tasks.
\newblock In \emph{Proceedings of the 18th Conference of the European Chapter of the Association for Computational Linguistics (Volume 2: Short Papers)}, pages 179--192, 2024.

\bibitem[Wu et~al.(2025)Wu, Ma, Liu, Zhang, Deng, Li, Chen, Zhang, Xue, and Plank]{wu2025m}
Chengyan Wu, Bolei Ma, Yihong Liu, Zheyu Zhang, Ningyuan Deng, Yanshu Li, Baolan Chen, Yi~Zhang, Yun Xue, and Barbara Plank.
\newblock M-absa: A multilingual dataset for aspect-based sentiment analysis.
\newblock \emph{arXiv preprint arXiv:2502.11824}, 2025.

\bibitem[Muhammad et~al.(2023)Muhammad, Abdulmumin, Ayele, Ousidhoum, Adelani, Yimam, Ahmad, Beloucif, Mohammad, Ruder, et~al.]{muhammad2023afrisenti}
Shamsuddeen~Hassan Muhammad, Idris Abdulmumin, Abinew~Ali Ayele, Nedjma Ousidhoum, David~Ifeoluwa Adelani, Seid~Muhie Yimam, Ibrahim~Sa'id Ahmad, Meriem Beloucif, Saif Mohammad, Sebastian Ruder, et~al.
\newblock {AfriSenti}: A twitter sentiment analysis benchmark for african languages.
\newblock In \emph{Proceedings of the 2023 Conference on Empirical Methods in Natural Language Processing}, pages 13968--13981, 2023.

\bibitem[Ibrahim et~al.(2025)Ibrahim, Zandam, Adam, Musa, Hassan, Hamada, and Usman]{ibrahim2025deep}
Umar Ibrahim, Abubakar~Yakubu Zandam, Fatima~Muhammad Adam, Aminu Musa, Mohamed Hassan, Mohamed Hamada, and Muhammad~Shamsu Usman.
\newblock A deep convolutional neural network-based model for aspect and polarity classification in hausa movie reviews.
\newblock \emph{Engineering Proceedings}, 107\penalty0 (1):\penalty0 21, 2025.

\bibitem[Maqsood et~al.(2025)Maqsood, Latif, and Latif]{maqsood2025towards}
Zoya Maqsood, Seemab Latif, and Rabia Latif.
\newblock Towards robust urdu aspect-based sentiment analysis through weakly-supervised annotation framework.
\newblock In \emph{Proceedings of the 8th International Conference on Natural Language and Speech Processing (ICNLSP-2025)}, pages 146--161, 2025.

\bibitem[Kaushik et~al.(2020)Kaushik, Hovy, and Lipton]{kaushik2020learning}
Divyansh Kaushik, Eduard Hovy, and Zachary~C Lipton.
\newblock Learning the difference that makes a difference with counterfactually-augmented data.
\newblock In \emph{International Conference on Learning Representations}, 2020.

\bibitem[{\v{S}}m{\'\i}d et~al.(2025){\v{S}}m{\'\i}d, P{\v{r}}ib{\'a}{\v{n}}, and Kr{\'a}l]{vsmid2025improving}
Jakub {\v{S}}m{\'\i}d, Pavel P{\v{r}}ib{\'a}{\v{n}}, and Pavel Kr{\'a}l.
\newblock Improving generative cross-lingual aspect-based sentiment analysis with constrained decoding.
\newblock \emph{arXiv preprint arXiv:2508.10369}, 2025.

\bibitem[Balashankar et~al.(2023)Balashankar, Subramanian, and Nelakuditi]{balashankar2023improving}
Ananth Balashankar, Lakshminarayanan Subramanian, and Srihari Nelakuditi.
\newblock Improving classifier robustness through active generative counterfactual data augmentation.
\newblock In \emph{Findings of the Association for Computational Linguistics: EMNLP 2023}, pages 1--14, 2023.

\bibitem[Wu et~al.(2024)Wu, Zhang, Zhou, and Xu]{wu2024diner}
Jialong Wu, Linhai Zhang, Deyu Zhou, and Guoqiang Xu.
\newblock Diner: Debiasing aspect-based sentiment analysis with multi-variable causal inference.
\newblock \emph{arXiv preprint arXiv:2403.01166}, 2024.

\bibitem[OpenAI(2024)]{openai_gpt4o_mini_2024}
OpenAI.
\newblock \textit{GPT-4o-mini} [large language model].
\newblock \url{https://platform.openai.com}, 2024.
\newblock Accessed: 2025-09-25.

\bibitem[Reimers and Gurevych(2019)]{reimers-2019-sentence-bert}
Nils Reimers and Iryna Gurevych.
\newblock Sentence-bert: Sentence embeddings using siamese bert-networks.
\newblock In \emph{Proceedings of the 2019 Conference on Empirical Methods in Natural Language Processing}. Association for Computational Linguistics, 11 2019.
\newblock URL \url{http://arxiv.org/abs/1908.10084}.

\bibitem[Wei and Zou(2019)]{wei2019eda}
Jason Wei and Kai Zou.
\newblock Eda: Easy data augmentation techniques for boosting performance on text classification tasks.
\newblock \emph{arXiv preprint arXiv:1901.11196}, 2019.

\bibitem[Schuster et~al.(2019)Schuster, Shah, Yeo, Filizzola, Santus, and Barzilay]{schuster2019debiasingfactverificationmodels}
Tal Schuster, Darsh~J Shah, Yun Jie~Serene Yeo, Daniel Filizzola, Enrico Santus, and Regina Barzilay.
\newblock Towards debiasing fact verification models, 2019.
\newblock URL \url{https://arxiv.org/abs/1908.05267}.

\bibitem[Niu et~al.(2021)Niu, Tang, Zhang, Lu, Hua, and Wen]{niu2021counterfactual}
Yulei Niu, Kaihua Tang, Hanwang Zhang, Zhiwu Lu, Xian-Sheng Hua, and Ji-Rong Wen.
\newblock Counterfactual vqa: A cause-effect look at language bias.
\newblock In \emph{Proceedings of the IEEE/CVF conference on computer vision and pattern recognition}, pages 12700--12710, 2021.

\bibitem[Liu et~al.(2021)Liu, Shen, He, Zhang, Xu, Yu, and Cui]{liu2021towards}
Jiashuo Liu, Zheyan Shen, Yue He, Xingxuan Zhang, Renzhe Xu, Han Yu, and Peng Cui.
\newblock Towards out-of-distribution generalization: A survey.
\newblock \emph{arXiv preprint arXiv:2108.13624}, 2021.

\bibitem[Conneau et~al.(2019)Conneau, Khandelwal, Goyal, Chaudhary, Wenzek, Guzm{\'a}n, Grave, Ott, Zettlemoyer, and Stoyanov]{conneau2019unsupervised}
Alexis Conneau, Kartikay Khandelwal, Naman Goyal, Vishrav Chaudhary, Guillaume Wenzek, Francisco Guzm{\'a}n, Edouard Grave, Myle Ott, Luke Zettlemoyer, and Veselin Stoyanov.
\newblock Unsupervised cross-lingual representation learning at scale.
\newblock \emph{arXiv preprint arXiv:1911.02116}, 2019.

\bibitem[Alabi et~al.(2022)Alabi, Adelani, Mosbach, and Klakow]{alabi-etal-2022-adapting}
Jesujoba~O. Alabi, David~Ifeoluwa Adelani, Marius Mosbach, and Dietrich Klakow.
\newblock Adapting pre-trained language models to {A}frican languages via multilingual adaptive fine-tuning.
\newblock In \emph{Proceedings of the 29th International Conference on Computational Linguistics}, pages 4336--4349, Gyeongju, Republic of Korea, October 2022. International Committee on Computational Linguistics.
\newblock URL \url{https://aclanthology.org/2022.coling-1.382}.

\bibitem[Adelani et~al.(2023)Adelani, Liu, Shen, Vassilyev, Alabi, Mao, Gao, and Lee]{adelani2023sib200}
David~Ifeoluwa Adelani, Hannah Liu, Xiaoyu Shen, Nikita Vassilyev, Jesujoba~O. Alabi, Yanke Mao, Haonan Gao, and Annie En-Shiun Lee.
\newblock Sib-200: A simple, inclusive, and big evaluation dataset for topic classification in 200+ languages and dialects, 2023.

\bibitem[Wang et~al.(2018)Wang, Mazumder, Liu, Zhou, and Chang]{wang2018target}
Shuai Wang, Sahisnu Mazumder, Bing Liu, Mianwei Zhou, and Yi~Chang.
\newblock Target-sensitive memory networks for aspect sentiment classification.
\newblock In \emph{Proceedings of the 56th Annual Meeting of the Association for Computational Linguistics (Volume 1: Long Papers)}, 2018.

\bibitem[Huang and Carley(2019)]{huang2019parameterized}
Binxuan Huang and Kathleen~M Carley.
\newblock Parameterized convolutional neural networks for aspect level sentiment classification.
\newblock \emph{arXiv preprint arXiv:1909.06276}, 2019.

\bibitem[Bai et~al.(2020)Bai, Liu, and Zhang]{bai2020investigating}
Xuefeng Bai, Pengbo Liu, and Yue Zhang.
\newblock Investigating typed syntactic dependencies for targeted sentiment classification using graph attention neural network.
\newblock \emph{IEEE/ACM Transactions on Audio, Speech, and Language Processing}, 29:\penalty0 503--514, 2020.

\end{thebibliography}

\appendix

\section{Appendix}
\label{sec:appendix}

\subsection{DINER FRAMEWORK}

\begin{figure}[h]
\centering
    \includegraphics[width=0.6\textwidth]{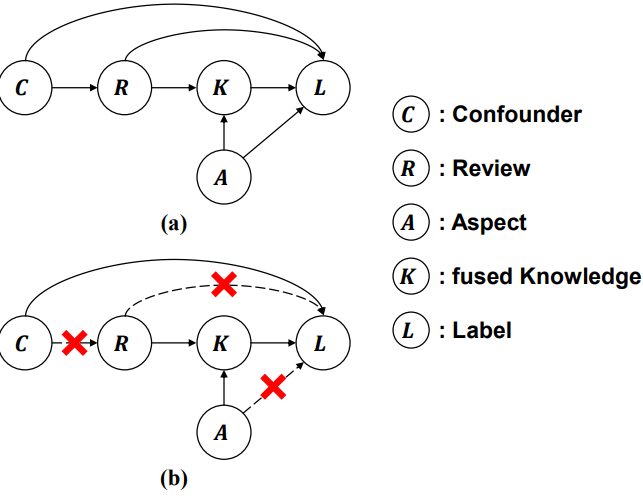}
    \caption{Causal structure of the DINER framework (\cite{wu2024diner}).}
    \label{fig:DAG}
\end{figure}

The DINER framework models ABSA as a Structural Causal Model (SCM) with five key variables: confounder $C$ (prior contextual knowledge), review $R$, aspect $A$, fused knowledge $K$ (derived from $R$ and $A$), and sentiment label $L$. The causal relations are defined as $R \rightarrow K \leftarrow A$, $K \rightarrow L$, $R \rightarrow L \leftarrow A$, and $C \rightarrow R, C \rightarrow L$. The causal structure is illustrated in Figure~\ref{fig:DAG}, showing the directed acyclic graph (DAG) that represents the relationships between these variables. The objective of DINER is to isolate the causal path $K \rightarrow L$ while eliminating spurious correlations through $R \rightarrow L \leftarrow A$ and minimizing confounding through $C \rightarrow R$ and $C \rightarrow L$. For details on causal graph construction, see \cite{wu2024diner}.  

\subsection{Causal Variables Discovery}
\begin{figure}[h]
    \centering
    \includegraphics[width=0.7\textwidth]{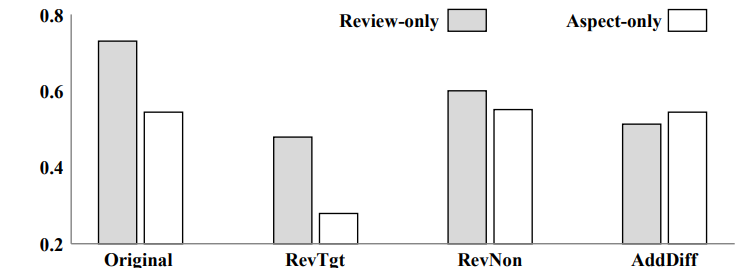}
    \caption{Demonstration of spurious correlations learned from annotation bias (\cite{wu2024diner}).}
    \label{fig:bias}
\end{figure}

Figure~\ref{fig:bias} demonstrates how high-capacity neural networks can easily learn spurious correlations arising from annotation bias. These biases cause models to capture misleading dependencies rather than true causal relationships. Due to their high representational power, neural models are particularly vulnerable to such biases, often learning shortcuts instead of underlying latent features (\cite{wang2018target,huang2019parameterized,bai2020investigating}). For instance, reversing the polarity of target aspects can result in up to a 20\% drop in accuracy, highlighting the model’s reliance on these correlations. The figure also supports the observation that models frequently depend on spurious correlations between target and non-target aspects in sentiment datasets such as \cite{marelli2014semeval}. In the “Original” setting, both review-only and aspect-only models perform well; however, when these correlations are disrupted in the “RevTgt,” “RevNon,” and “AddDiff” settings, performance decreases or becomes balanced. This confirms that neural models often exploit dataset biases rather than learning genuine target-specific sentiment representations.

\begin{table*}[t]
\centering
\small
\setlength{\tabcolsep}{4pt}
\renewcommand{\arraystretch}{1.0}
\begin{tabular}{p{1.5cm}p{9.5cm}ccc}
\toprule
\textbf{Type} & \textbf{Examples (Target Aspect: food)} & \textbf{Gold} & \textbf{Baseline} & \textbf{DINER} \\
\midrule
Original & The food is top notch, the service is attentive, and the atmosphere is great. & Positive & Positive \checkmark & Positive \checkmark \\
RevTgt & The food is nasty, but the service is attentive, and the atmosphere is great. & Negative & Negative \checkmark & Negative \checkmark \\
RevNon & The food is top notch, the service is heedless, but the atmosphere is not great. & Positive & Negative \ding{55} & Positive \checkmark \\
AddDiff & The food is top notch, the service is attentive, and the atmosphere is great, but the music is too heavy and the staff is arrogant. & Positive & Negative \ding{55} & Positive \checkmark \\
\bottomrule
\end{tabular}
\caption{Illustrative case studies showing model robustness under target and non-target sentiment perturbations.}
\label{tab:case_studies}
\end{table*}
To mitigate the indirect confounding between sentiment words in the review and the context, DINER employs a backdoor adjustment intervention that addresses the $C \rightarrow R \rightarrow L$ pathway. The intervention uses the do-calculus formulation:

\begin{equation}
\begin{split}
P(L|do(R)) &= \sum_c P(L|R,C)P(C) \\
&= \sum_c \frac{P(L,R|C)P(C)}{P(R|C)}
\end{split}
\end{equation}

The debiased review representation is computed as:

\begin{equation}
\begin{split}
\zeta_{r'} = \frac{\tau}{K} \sum_k &\frac{(w_k)^T}{(||w_k||+\epsilon)} \cdot \\
&\left(\frac{r_k}{||r_k||} - \frac{r_{kc}}{||r_{kc}||}\right)
\end{split}
\end{equation}

To address the direct correlation between aspect and label ($A \rightarrow L$), DINER employs counterfactual reasoning using the Total Indirect Effect (TIE):

\begin{equation}
\begin{split}
TIE_{a,r} &= TE_{a,r} - NDE_r - NDE_a + IE_{a,r} \\
&= TE_{a,r} - (NDE_r + NDE_a) \\
&\quad (\text{since } IE_{a,r} = 0)
\end{split}
\end{equation}

Following \cite{niu2021counterfactual}, we define:
\begin{equation}
\begin{split}
TE_{a,r}  &= L_{a,r,k} - L_{a^*,r^*,k^*} \\
NDE_a     &= L_{a,r^*,k^*} - L_{a^*,r^*,k^*} \\
NDE_r     &= L_{a^*,r,k^*} - L_{a^*,r^*,k^*}
\end{split}
\end{equation}

Thus:
\begin{equation}
\begin{split}
TIE_{a,r} = L_{a,r,k} - L_{a^*,r,k^*} \\
- L_{a,r^*,k^*} + L_{a^*,r^*,k^*}
\end{split}
\end{equation}

The final prediction is implemented through a model ensemble (\cite{wu2024diner}):

\begin{equation}
\begin{split}
L_{a,r,k} &= L(A\!=\!a, R\!=\!r', K\!=\!k) \\
&= \Psi(\zeta_a, \zeta_{r'}, \zeta_k) \\
&= \zeta_k \!+\! \tanh(\zeta_a) \!+\! \tanh(\zeta_{r'})
\end{split}
\end{equation}

Where $\zeta_{r'}$ is the output of the debiased review-only branch, $\zeta_a$ is the output of the aspect-only branch, and $\zeta_k$ is the output of the fused knowledge branch.

\subsection{Case Studies}
\label{sec:case_study}
Table~\ref{tab:case_studies} presents illustrative examples demonstrating the robustness improvements from our causal training. In the original example, both baseline and DINER correctly classify positive sentiment. For RevTgt cases, both models correctly adjust when target sentiment is reversed. However, for RevNon scenarios, DINER correctly maintains focus on target aspect despite negative non-target aspects, while baseline models fail. Similarly, with AddDiff examples, DINER shows robustness when additional negative aspects are introduced. These examples highlight DINER's ability to isolate target aspect sentiment from confounding context—a crucial capability for practical ABSA applications.

\end{document}